\documentclass{article}

\usepackage[final]{neurips_2024}
\usepackage{graphicx}
\usepackage{multirow}
\usepackage{xcolor}
\usepackage{framed}

\usepackage{natbib} 

\usepackage[utf8]{inputenc} 
\usepackage[T1]{fontenc}    
\usepackage{hyperref}       
\usepackage{url}            
\usepackage{booktabs}       
\usepackage{amsfonts}       
\usepackage{nicefrac}       
\usepackage{microtype}      
\usepackage{xcolor}         
\usepackage{amsmath}
\usepackage{natbib}

\title{Enhancing Table Representations with LLM-powered Synthetic Data Generation}

\author{%
  Dayu Yang\thanks{Work done when interning at Capital One} \\
  University of Delaware\\
  \texttt{dayu@udel.edu} \\
  \And
  Natawut Monaikul \\
  Capital One \\
  \texttt{natawut.monaikul@capitalone.com} \\
  \AND
  Amanda Ding \\
  Capital One \\
  \texttt{amanda.ding@capitalone.com} \\
  \And
  Bozhao Tan \\
  Capital One \\
  \texttt{bozhao.tan@capitalone.com} \\
  \And
  Kishore Mosaliganti \\
  Capital One \\
  \texttt{kishore.mosaliganti@capitalone.com}\\
  \And
  Giri Iyengar \\
  Capital One \\
  \texttt{giridharan.iyengar@capitalone.com} \\
}

\begin{document}

\maketitle

\begin{abstract}

In the era of data-driven decision-making, accurate table-level representations and efficient table recommendation systems are becoming increasingly crucial for improving table management, discovery, and analysis. However, existing approaches to tabular data representation often face limitations, primarily due to their focus on cell-level tasks and the lack of high-quality training data. To address these challenges, we first formulate a clear definition of table similarity in the context of data transformation activities within data-driven enterprises. This definition serves as the foundation for synthetic data generation, which require a well-defined data generation process. Building on this, we propose a novel synthetic data generation pipeline that harnesses the code generation and data manipulation capabilities of Large Language Models (LLMs) to create a large-scale synthetic dataset tailored for table-level representation learning. Through manual validation and performance comparisons on the table recommendation task, we demonstrate that the synthetic data generated by our pipeline aligns with our proposed definition of table similarity and significantly enhances table representations, leading to improved recommendation performance.

\end{abstract}

\section{Introduction}

In an era where data-driven decision-making is increasingly central to business operations, the ability to efficiently and accurately recommend similar tables across vast datasets has become essential~\citep{zhang_ad_2018, habibi_tabsim_2020, trabelsi_strubert_2022}. As tables remain a dominant modality in the data landscape, this capability can greatly enhance both table management and discovery processes~\citep{habibi_tabsim_2020, deng2022turl}. By recommending similar tables, organizations can streamline table management through effective deduplication, precise lineage prediction, and robust clustering or labeling~\citep{yin_tabert_2020, iida_tabbie_2021}. These processes contribute to maintaining clean, organized, and well-documented data repositories, which can lead to significant cost savings in cloud services~\citep{mahesh2020review}. Furthermore, similar table recommendations play a crucial role in discovery and usage, suggesting complementary tables that offer additional insights~\citep{zhang_ad_2018}. This enables data analysts to make more informed decisions and better monitor ongoing processes.

Although similar table recommendation is a crucial task, many existing table similarity methods~\citep{zhang_ad_2018, habibi_tabsim_2020, trabelsi_strubert_2022} often lack a clear and consistent definition of "similarity." This ambiguity can make it challenging to apply these methods to different use cases, as users may be unsure whether their understanding of similarity aligns with the definitions used in these approaches.



A single table can contain a large amount of cell data and have a complex structure, making the manual annotation of similar tables a costly process. This contributes to the scarcity of high-quality training data. Facing the issue, some studies~\citep{yin_tabert_2020, iida_tabbie_2021, chen_hytrel_2023} have attempted to address the challenge by developing table representations through unsupervised cell-level tasks, such as masked cell value restoration. However, these representations tend to struggle with capturing the global structure of a table, which may result in suboptimal performance in table-level tasks, including similar table recommendation, as suggested by our experimental results. Another line of research has sought to alleviate the data sparsity challenge~\citep{habibi_tabsim_2020, trabelsi_strubert_2022} by reframing the table recommendation problem as a \emph{pairwise} table matching task, rather than focusing on \emph{pointwise} table representation. This approach is generally easier to solve and requires less supervision~\citep{melnikov2016pairwise, yates2021pretrained}. However, while pairwise tasks may mitigate some data limitations, they introduce significant computational inefficiencies, shifting the time complexity from linear to quadratic~\citep{melnikov2016pairwise, zhuang2024setwise}, which can make them impractical for large-scale tabular databases typically found in modern big data environments.

To address the limitations of existing methods in table similarity recommendation, we propose a structured approach that begins with a clear definition of table similarity grounded in real-world use cases from data-driven industries. Building upon this foundation, we develop a synthetic table similarity data generation pipeline leveraging Large Language Models (LLMs). By providing any non-annotated tables as a base, this pipeline can generate large-scale, high-quality table similarity data for enhancing general table representation model and recommendation performance evaluation.


We evaluated the quality of our generated synthetic dataset through three approaches: (i) conducting human validation on a subset of the dataset, confirming its accuracy with respect to our defined notion of similarity; (ii) comparing cosine similarities of embeddings from similar tables in our dataset against those from an existing dataset, demonstrating its enhanced potential for developing table-level representations; and (iii) improving table representations using our synthetic dataset for the task of similar table matching, outperforming state-of-the-art embedding models in our out-of-distribution proprietary dataset collected from Capital One internally. These results indicate that our method can generate high-quality synthetic training data and contribute to real-world similar table recommendation applications.

\section{Related Studies}

\subsection{Textual Representation Learning}

Text embeddings -- vector representations of natural language that capture semantic content -- have been widely utilized in various natural language processing (NLP) tasks, including question answering~\citep{choi2018quac, allam2012question}, conversational search~\citep{yang2023exploration, yang2023zero}, and semantic textual similarity~\citep{muennighoff2022mteb}. Sentence-BERT~\citep{reimers_sentence-bert_2019} is one of the earliest and most popular methods for learning text embeddings by fine-tuning BERT~\citep{devlin2018bert} on natural language inference datasets. 

To further improve the performance and robustness of text embeddings, state-of-the-art methods like GTE-series~\citep{li2023gte} and BGE-series~\citep{bge_embedding} adopt a more complex multi-stage training approach. These methods first pre-train on billions of weakly supervised text pairs and then fine-tune on several high-quality labeled datasets.

Recently, with the growing recognition of the strong language understanding capabilities of large language models (LLMs), there has been a shift towards using LLMs, particularly those based on decoder Transformers, to generate text embeddings~\citep{lee2024nvembed,behnamghader2024llmvec}. In addition to changes in model structure, recent state-of-the-art text representation models have begun leveraging larger LLMs to generate synthetic data, providing high-quality supervision, combined with large-scale weak supervision from traditional retrieval datasets like MSMARCO~\citep{wang_improving_2024, lee_gecko_2024}. The success of synthetic data generation techniques on text representation learning indicates its great potential on enhancing table representation.

\subsection{Tabular Representation Learning}

Inspired by the success of BERT~\citep{devlin2018bert} in constructing general text representations and achieving strong performance across various downstream natural language processing tasks, many studies in table representation~\citep{yin_tabert_2020, iida_tabbie_2021, chen_hytrel_2023} have adopted a similar training approach -- they aim to build general table representations using masked self-supervised tasks. Specifically, they modify the vanilla Transformer encoder~\citep{vaswani_attention_2017} by introducing additional column or row attention mechanisms to better adapt the model to learn the spatial structure of a 2D table. In these works, a large unannotated tabular dataset is used for pre-training, where approximately 15\% of the cells are masked, and the modified Transformer is trained to predict these masked cells.

Recognizing the success of LLMs such as Llama~\citep{touvron2023llama} and Gemma~\citep{team2024gemma} in natural language tasks, some studies have explored their potential applications in tabular data tasks, such as tabular data cleaning, cell value lookup, and tabular data classification~\citep{borisov2022deep}. To apply LLMs directly to tabular data, the table must first be serialized into a natural text format. However, it remains an open question as to which format -— CSV, JSON, XML, Markdown, or HTML~\citep{sui2024table}.

Other methods have been developed to directly address the task of estimating table similarity or performing table retrieval~\citep{habibi_tabsim_2020,trabelsi_strubert_2022}. These methods take a pair of tables simultaneously, giving a \emph{pairwise} representation. However, we seek a \emph{pointwise} representation -- a unique table-level embedding -- that can allow for more efficient table retrieval as well as be used for other downstream tasks involving tabular data.

\subsection{Data for Table Similarity Estimation}

Currently, only a few datasets for table similarity have been proposed in the literature. One such dataset uses tables from PubMed Central (PMC) Open Access -- pairs of tables were manually annotated as similar or dissimilar according to the estimated percentage of similarity or dissimilarity among the cell data and captions in the tables~\citep{habibi_tabsim_2020, trabelsi_strubert_2022}. These tables were drawn from a predominantly biomedical and scientific domain. Another more domain-general dataset draws tables from Wikipedia, and table pairs were manually annotated for equivalence or subset relationships based on matching column names~\citep{zhang_ad_2018, fetahu_tablenet_2019, habibi_tabsim_2020}. While these datasets are the most relevant to our work, the annotation guidelines in both reflect limited definitions of similarity.

Other datasets for table similarity have been adapted from datasets for table retrieval -- these datasets, such as WikiTables~\citep{bhagavatula2015tabel,zhang_ad_2018} and TableArXiv~\citep{gao_scientific_2017}, were originally developed for retrieving relevant tables given a natural-language query. Studies on table similarity then define a pair of tables as similar if both have been labeled as relevant to the same query~\citep{habibi_tabsim_2020,trabelsi_strubert_2022}. While this definition is useful for grouping tables around shared keywords in a query, relevance judgments are not between the tables themselves, so the sense of table similarity is less targeted.


We also note that in the performance evaluation of StruBERT~\citep{trabelsi_strubert_2022}, a state-of-the-art model for predicting if a pair of tables is similar, we found a sort of label leakage in the test set. Although the test set does not contain pairs also present in the training set, the individual tables overlap between the sets. Because of this, about 74\% of the pairs in the test set could be inferred transitively from relationships in the training set -- for example, if the model learns that tables A and B are similar, and B and C are similar from the training set, it could infer that A and C are similar in the test set.


We removed these label-leakage pairs from both the training and test sets and trained the StruBERT model, as well as a vanilla BERT (base) model using a na\"{i}ve table serialization scheme, on the new training set. We found the adjusted performance of StruBERT that we calculated on the new test set to have an accuracy of 0.874, compared to 0.9942 reported on the original test set. This ends up underperforming compared to the vanilla BERT model (an accuracy of 0.908). Furthermore, we found the average latency incurred by StruBERT performing inference on each sample to be around five times longer than BERT due to the added complexity of the model.

Taken together, our literature review demonstrates the need for a large-scale, domain-general dataset of table pairs that are labeled according to a practical definition of similarity to enable tabular representation learning and to efficiently perform and fairly evaluate the task of similar table retrieval in data-driven industries.

\section{Definition of Similarity}

In our work, we define “similarity” based on two key use cases of table matching systems in industry: table management and complementary information retrieval. In industry, table management systems are often designed to identify duplicate tables and those with close lineage. While exact duplicates can be easily identified through hard-coded rule matching, finding tables with close lineage is more challenging. This is because data analysts often modify or transform elements of parent tables, requiring the model to understand the underlying semantic connections among tables.

Another critical use case in table recommendation is the retrieval of complementary information. In this context, the goal is to identify tables that, while not identical, offer additional insights or relevant data that can enhance the analysis. This requires the model to recognize nuanced relationships between tables, such as shared themes, overlapping data points, or related metadata.

Taking these two aspects into account, we define two tables to be \emph{similar} if one is the result of the other's having undergone one or more data transformations typically performed by a data analyst. We elaborate further on these types of operations in the next section. This definition captures the semantic connections and transformations that occur in real-world data management, ideally resulting in table-level representations that allow for effective identification and recommendation of other contextually-related tables.

\section{Synthetic Data Generation Pipeline}
\label{sec:pipeline}

We now introduce our LLM-assisted pipeline for generating tabular data from a given input table such that the generated table is considered similar to the original table, i.e., the generated table can be obtained from some series of transformations on the anchor table.

We first assume we are given a table, which we call an \emph{anchor} table, from which our goal is to generate similar tables in a way that (1) mimics the behavior of human analysts, (2) encompasses a diverse range of data operations, (3) is efficient. We also assume that this anchor table minimally contains a title, column names, some cell data, and some sort of description that briefly summarizes the contents and purpose of the table.


We then perform one or more tabular data operations on the anchor table to generate a new table that is similar to the anchor table. We manually constructed a concise list of possible table transformations informed by a study of an industry documentation on data management for data analysts:

\begin{itemize}
    \item \textbf{Concatenation}: Add one or more new columns with new and relevant information.
    \item \textbf{Edit}: Create a new column based entirely on an existing column, using string operations like regular expressions, information extraction, or information refinement.
    \item \textbf{Reordering}: Shuffle the order of columns.
    \item \textbf{Calculation}: Generate a calculated column based on an existing numerical column.
    \item \textbf{Removal}: Remove one or more columns.
    \item \textbf{Update}: Modify the title, description, and column names with respect to any new values, or simply to re-word.
\end{itemize}

An example of these operations performed on a sample table is given in Appendix~\ref{sec:operations}. We believe these operations cover three major aspects of what a data analyst typically performs on the job: adding information (Concatenation), deleting information (Removal), and modifying or synthesizing information (Edit, Reordering, Calculation, Update). The reordering and update operations also help to ensure the generated tables can promote robust tabular embeddings in terms of order invariance and semantic representation.

Whereas the reordering and removal operations can be performed programmatically, we call on an LLM to perform the remaining, more complex operations. After choosing a set of table transformations to perform on the anchor table, each transformation is applied sequentially (i.e., to the output of the preceding transformation) -- we found that if an LLM is given multiple operations in a single prompt, the LLM frequently fails to apply \emph{all} of the operations.

We do not include any row operations because, compared to columns, each row represents an observation, and the number of observations typically exceeds the number of columns (or variables). When tables are serialized for input into a language model, only a few rows are typically sampled~\citep{yin_tabert_2020, trabelsi_strubert_2022, sui2024table}. Therefore, we eschew spending significant computation time on row operations.

The output of this pipeline is then multiple tables that can be considered similar to the anchor table, where each output table is the result of some set of tabular transformations applied to the anchor table that mirror those that a data analyst would apply to tabular data in practice. Then, given a sizable pool of anchor tables, we can generate a large-scale synthetic dataset of pairs of similar tables that can then be used to train and evaluate embedding models and models for downstream tasks involving tabular data.

\section{Evaluation}
To empirically verify the validity and utility of our pipeline, we evaluate the output of the pipeline on a standard set of anchor tables via manual inspection, an investigation of similarity with respect to embeddings, and a performance comparison on the downstream task of table retrieval.

In our pipeline implementation, we use the \texttt{Llama3.1-8B-Instruct} model as the LLM for performing tabular transformations. For each anchor table, two random sets of operations are chosen to perform on the table, generating two similar tables per anchor. Practical considerations for the order of operations are detailed in Appendix~\ref{sec:operations}.

We drew anchor tables from the WikiTables dataset~\citep{bhagavatula2015tabel}, which we chose in consideration of its size (approximately 1.6 million tables) and broad coverage of diverse topics. Each table in the dataset contains a title, column names, and cell data -- because our pipeline also expects a description for each table, we additionally prompt the same LLM to generate a brief description for about 700,000 randomly-chosen tables.

The exact prompts we designed to have the LLM perform each operation is given in Appendix~\ref{sec:prompts}. Each table in the dataset is represented in a JSON format, and we serialize these tables for the LLM by directly taking the string representation of the JSON table -- details are given in Appendix~\ref{sec:serialization}. We prompt the LLM to also produce JSON-formatted tables, although the output was not always in a valid JSON format, so we only save pairs for which the outputs could be properly parsed. The generated dataset output from our pipeline ultimately contained 140,000 pairs of tables (70,000 anchor tables with 2 synthetic similar tables each).\footnote{We also plan to release our generated dataset pending approvals.}

\subsection{Manual Validation}
An immediate limitation of our pipeline is in the prompt engineering required to ensure the LLM produces sensible tables that truly reflect the instructions in the prompt. While it is possible to use the LLM itself to generate a list of operations or sufficient prompts to instruct itself to product new tables, similar to other studies in LLM-based synthetic data generation~\citep{wang_improving_2024,lee_gecko_2024}, this approach is inefficient because the list of operations generated by the LLM can be long and contain significant overlap. Given that we apply operations sequentially, this would entail calling on the LLM many more times, which would incur a significant computational cost.

To verify that our LLM prompting generated reasonable tables, we randomly selected a subsample of the generated dataset for a round of manual validation. Two professional data analysts inspected each pair of tables (an anchor table and its corresponding generated table) to determine if the operations that were performed by the LLM were indeed correct. Though multiple operations could be performed on an anchor table, the annotators only saw the anchor table and the final generated table; thus, the annotators were asked to label simply whether or not all of the listed operations had been performed correctly.

We randomly selected 80 table pairs for manual review. The data analysts independently labeled these samples, after which disagreements were discussed to arrive at a final label. We measured the inter-annotator agreement using Cohen's kappa~\citep{cohen_coefficient_1960} on the initial labels and found moderately strong agreement between the annotators (\( \kappa = 0.56 \))~\citep{landis1977measurement}. Our manual review found that 65\% of the synthetic tables were generated correctly given their respective sets of operations, with a majority of the incorrect tables being due to the edit operation. These results suggest that our proposed pipeline is able to output tables that reflect typical tabular transformations in data-driven industries, but that further prompt engineering may be required to improve the pipeline's abilities on more complex transformations.

\subsection{Embedding-Based Similarity Validation}
Next, we investigate the potential for our generated dataset to be used to build a robust tabular representation model. As is typical for representation learning, we envision an embedding space for tabular data in which the vector representations of similar tables are close to each other. Specifically, the similarity of two tables \( T_1 \) and \( T_2 \) should be a function of the similarity of their embeddings \( E_1 \) and \( E_2 \) given some embedding model.

We compared the \emph{cosine similarities} of embeddings of pairs of similar tables in our generated dataset against those pairs of similar tables in the dataset used to train and evaluate StruBERT~\citep{trabelsi_strubert_2022}, which is also based on the WikiTables dataset. We seek to show that with our targeted definition of similarity and our synthetic data generation pipeline developed around that definition, the pairs of tables in our generated dataset exhibit relatively high similarity scores.

For a fair comparison, we do not train an embedding model on our generated dataset; rather, we use a state-of-the-art text embedding model -- \texttt{bge-large-en-v1.5} -- to embed the tables, from which we can compute cosine similarities of pairs of tables:
\[
\text{cosine similarity}(T_1,T_2) = \frac{E_1 \cdot E_2}{\|E_1\| \|E_2\|}
\]

We serialized a table by simply concatenating its title, columns names, and the cell data of a randomly-selected row, separated by periods and commas (between cells). The row to include was selected randomly so that the generated table pairs are not given an advantage when the first row of both tables are mostly the same. Descriptions were not included since the original WikiTables dataset does not contain them. 

\begin{figure}
    \centering
    \includegraphics[width=0.4\linewidth]{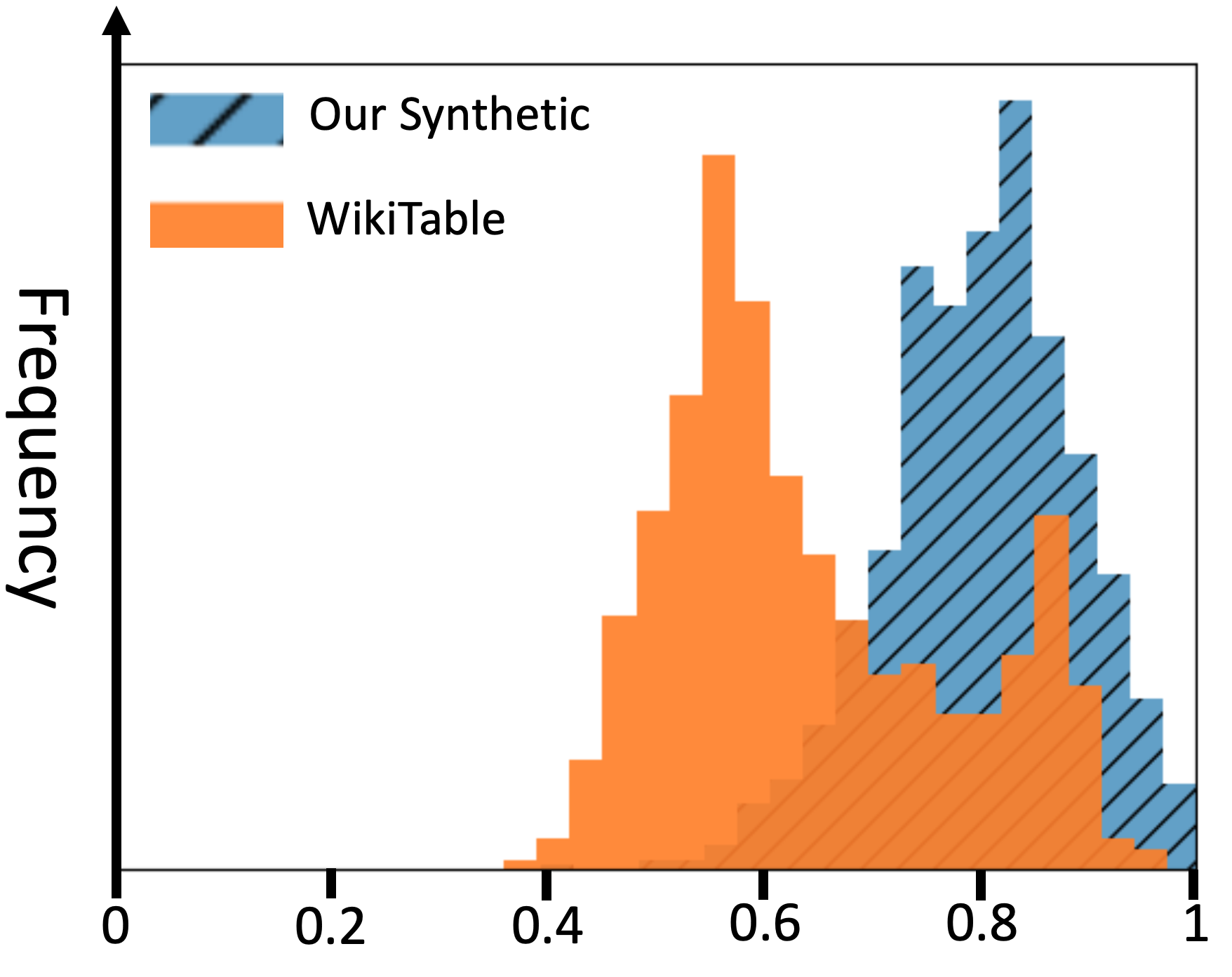}
    \caption{Cosine similarity scores of BGE-based embeddings of 1,000 pairs of similar tables, from our synthetic dataset and from the labeled portion of the WikiTables dataset used for StruBERT.}
    \label{fig:bge-dist}
\end{figure}

Figure~\ref{fig:bge-dist} shows the distribution of cosine similarity scores of 1,000 randomly-sampled similar table pairs from our generated dataset and 1,000 randomly-sampled similar table pairs from the dataset used for StruBERT. We see that cosine similarity scores of our generated dataset center around a higher value, demonstrating that our definition of similarity and our data generation pipeline can produce a dataset of pairs of similar tables such that the similarity is reflected in an embedding space. Since we only used a na\"{i}ve serialization and a pre-trained embedding model, this also suggests that our dataset can serve as a basis for building a robust table-level representation model.

\subsection{Downstream Validation}
Finally, we evaluate the quality of our generated dataset by its utility for a downstream task -- specifically, similar table retrieval. An embedding model fine-tuned on a high-quality dataset of pairs of similar tables should be able to perform better on a similar table retrieval task than other embedding models, since the fine-tuned model will have learned better representations of tabular data for determining similarity.

In the similar table retrieval task, given a query table \( T_q \), the objective is to find similar tables from a corpus of tables \( C = \{T_1, T_2, \dots, T_k\} \), where \( k \) is the total number of tables. A typical solution for this task would be to use an embedding model to take \( T_q \) as input and output an embedding \( E_q \) of \( T_q \). The cosine similarity can then be computed between \( E_q \) and the embeddings of all other tables; those with the highest similarity scores would be considered candidate similar tables.

We randomly split our generated dataset based on 70,000 anchor tables into a training set containing 55,000 anchors, a validation set containing 5,000 anchors, and a test set containing 10,000 anchors, each with their generated pairs. Given that we generate two similar tables per anchor table, the task then for a representation model is to generate representations that enable the use of cosine distance to rank the two ground truth similar tables as high as possible from a pool of around 700,000 tables for each of the anchor tables in the test set. Note that the pool then is made up of original WikiTables data as well as the generated tables corresponding to the anchor tables in the test set.

\subsubsection{Baseline}

We first evaluate state-of-the-art embedding models on this task directly to establish a baseline of pre-trained model performance. We compare three models: \texttt{bge-large-en-v1.5} (Short as BGE), \texttt{E5-mistral-7b-instruct}~\citep{wang_improving_2024} (a more powerful representation model with about 24x more parameters than bge-large-en-v1.5), and TABBIE~\citep{iida_tabbie_2021} (a general table representation model). For the two text embedding models, we again serialize tables as described in Appendix~\ref{sec:serialization}. For TABBIE, we use the \texttt{[CLS]} embedding at the \( (0,0) \) position as the representation for the entire table, as the original authors had also done in their experiments. Since TABBIE does not take into account a table's title or description, to make a fairer comparison, we also embed each table's concatenated title and description using \texttt{bge-large-en-v1.5} and compute similarity as a weighted average between the similarity of the text embeddings and the similarity of the TABBIE embeddings.\footnote{We choose to give 90\% of the weight to the TABBIE embeddings in order to include the title and description embeddings but to primarily focus on the TABBIE embeddings.} We refer to this method as \emph{TABBIE-plus}.




As metrics, we compute two standard measures of retrieval and ranking performance: recall and nDCG. In addition to computing these metrics in the top ten returned results, since each anchor table has exactly two similar tables in the pool, we also compute recall and nDCG in only the top two returned results. These results are given in Table~\ref{tab:retrieve_testing}. We see that the BGE model outperforms the other models, despite being smaller and not originally intended for tabular data. Therefore, we chose BGE model as the base model that further implemented for fine-tuning on our synthetic data. 

\begin{table}[]

\caption{Performance comparison of the pointwise table representation models on synthetic data.}
\centering
\begin{tabular}{l|cccc}

\hline
                       & Recall@2        & nDCG@2          & Recall@10       & nDCG@10         \\ \hline
bge-large-en-v1.5      & 0.8037          & 0.9232          & 0.8880          & 0.9253          \\
E5-mistral-7b-instruct & 0.7830           & 0.9000          & 0.9073          & 0.9097          \\
TABBIE-plus            & 0.6609 & 0.6889 & 0.7902 & 0.7545 \\ \hline
\end{tabular}
\label{tab:retrieve_testing}
\end{table}


\subsubsection{Fine-Tuning}

With this baseline, we seek to show that fine-tuning the BGE model on our synthetic dataset will further refine its representations of tabular data such that it will not only perform even better on this test set, but also  be able to generalize to a completely different test set of close-lineage tables for an industry use case.

We fine-tune the BGE model using contrastive learning. To achieve this, each anchor table and its corresponding similar tables must be accompanied by some hard negative examples -- tables that are dissimilar to the anchor table and should therefore have representations that are farther away from the anchor table. For each anchor table, hard negatives were sourced from the pool of 700,000 tables through a hybrid-search pipeline using both semantic and bag-of-words embeddings.

Given a pair of similar tables -— an anchor table \( T_a \) and a generated target table \( T_t \) -— the cosine similarity between their embeddings should be relatively higher than the cosine similarity between \( T_a \) and a dissimilar table or hard negative. For fine-tuning, we apply the standard InfoNCE loss \( L \) over hard negatives and in-batch negatives:

\[
L = - \log \frac{\exp(\phi(T_a, T_t))}{\sum_{T_n \in N} \exp(\phi(T_a, T_n))}
\]

where \( N \) denotes the set of all negatives, and $T_n$ represent a dissimilar table sampled from \( N \). \( \phi \) is the same temperature-scaled cosine similarity defined above. 

For training, we set the batch size to 4. Although contrastive learning typically benefits from larger batch sizes, we limited the batch size to 4 to fit within the memory constraints of a single GPU. The maximum serialized length for all input tables was set to 512 tokens, truncating any table exceeding this length. For each sample, we used 15 hard negatives in addition to the in-batch negatives.

In addition to evaluating this fine-tuned model on the test set from our generated dataset, we also measure its performance on a proprietary dataset from our industry. This dataset contains proprietary tabular data that have been labeled for close-lineage relationships, i.e., if one table was produced from another table through some data transformation, making it suitable for evaluating the fine-tuned model on similar table retrieval. While the Wikipedia tables are more general and easily understandable, the tables in this proprietary dataset are more specialized, containing domain-specific terminology. As a result, this dataset is considered out-of-distribution for the model, making it a crucial benchmark for assessing the model's value in real-world applications.

This dataset consists of about 8,000 tables, from which we sample 1,000 to perform our evaluation. The task then is for the model to generate representations that allow the similar (close-lineage) tables to rank as highly as possible among the pool of 8,000 tables for each of the tables in the evaluation. We also note that each table contains a title, description, and column names, but no cell data for privacy purposes; thus, in serializing the tables, we simply leave the cell data blank.

\begin{table}[]

\caption{Performance comparison of the BGE model before and after fine-tuning on synthetic data, evaluated on synthetic data and proprietary out-of-distribution data.}
\centering
\begin{tabular}{l|l|cccc}

\hline
Dataset &     Model         & Recall@2        & NDCG@2          & Recall@10       & NDCG@10         \\ \hline
Synthetic & BGE      & 0.8037          & 0.9232          & 0.8880          & 0.9253          \\
& Ours (BGE + Synthetic Data)             & \textbf{0.9043} & \textbf{0.9779} & \textbf{0.9329} & \textbf{0.9768} \\ \hline
\hline
Industry & BGE      & 0.3972          & 0.3062          & 0.5781          & 0.3126          \\
& Ours (BGE + Synthetic Data)             & \textbf{0.4662} & \textbf{0.3517} & \textbf{0.6681} & \textbf{0.3539} \\ \hline
\end{tabular}
\label{tab:retrieve_testing2}
\end{table}

Table~\ref{tab:retrieve_testing2} shows the improved performance of the fine-tuned model on the synthetic dataset -- this result is expected, given that the test set shares a similar distribution and similar features as the training set. The more challenging evaluation involves the proprietary dataset of similar tables, which contains out-of-distribution samples. As also shown in Table~\ref{tab:retrieve_testing2}, the fine-tuned model still improves upon the pre-trained BGE model, despite being fine-tuned on synthetic data that is quite different from the proprietary dataset. This demonstrates that the data generated from our pipeline allows for more robust tabular data representations, enhancing table similarity retrieval performance even for out-of-distribution samples.

\section{Conclusion}

In this paper, we enhanced table-level representation for similar table recommendation tasks using large language models (LLMs). We identified two key challenges in the field -— data sparsity and the ambiguous definition of table similarity -— and alleviated them by introducing a novel synthetic data generation process based on LLMs and clearly defining the table similarity problem. We then demonstrated the quality and utility of our generated dataset through manual validation, comparing embeddings to an existing table similarity dataset, and evaluating models using our data on the downstream task of similar table recommendation.

Our evaluations, conducted on both synthetic and proprietary datasets, comprehensively demonstrate that the proposed method effectively improves table similarity matching, even in scenarios involving out-of-distribution samples. The results suggest that our approach has the potential to bridge the gap between synthetic training data and practical applications, offering a viable solution for similar table recommendation in data-driven environments.

While these findings are promising, further research is needed to explore the scalability of our method across even larger datasets. Additionally, improving the ability of LLMs to generate desired and complete JSON-formatted tables remains a crucial area for future work.

\bibliographystyle{plainnat}
\bibliography{C1}

\appendix
\section{Appendix}

\subsection{Table Serialization}
\label{sec:serialization}

Our tables are originally represented as JSON objects. When serializing a JSON object to text that fits the input format of a language model, we directly convert the JSON object into a string, following the order: cell data, description, title, and column names. For cell data, we randomly sample only two observations to avoid exceeding the context limit of the language model.

\subsection{Tabular Operations}
\begin{figure}
    \centering
    \includegraphics[width=\linewidth, trim={5cm 15cm 5cm 15cm}, clip]{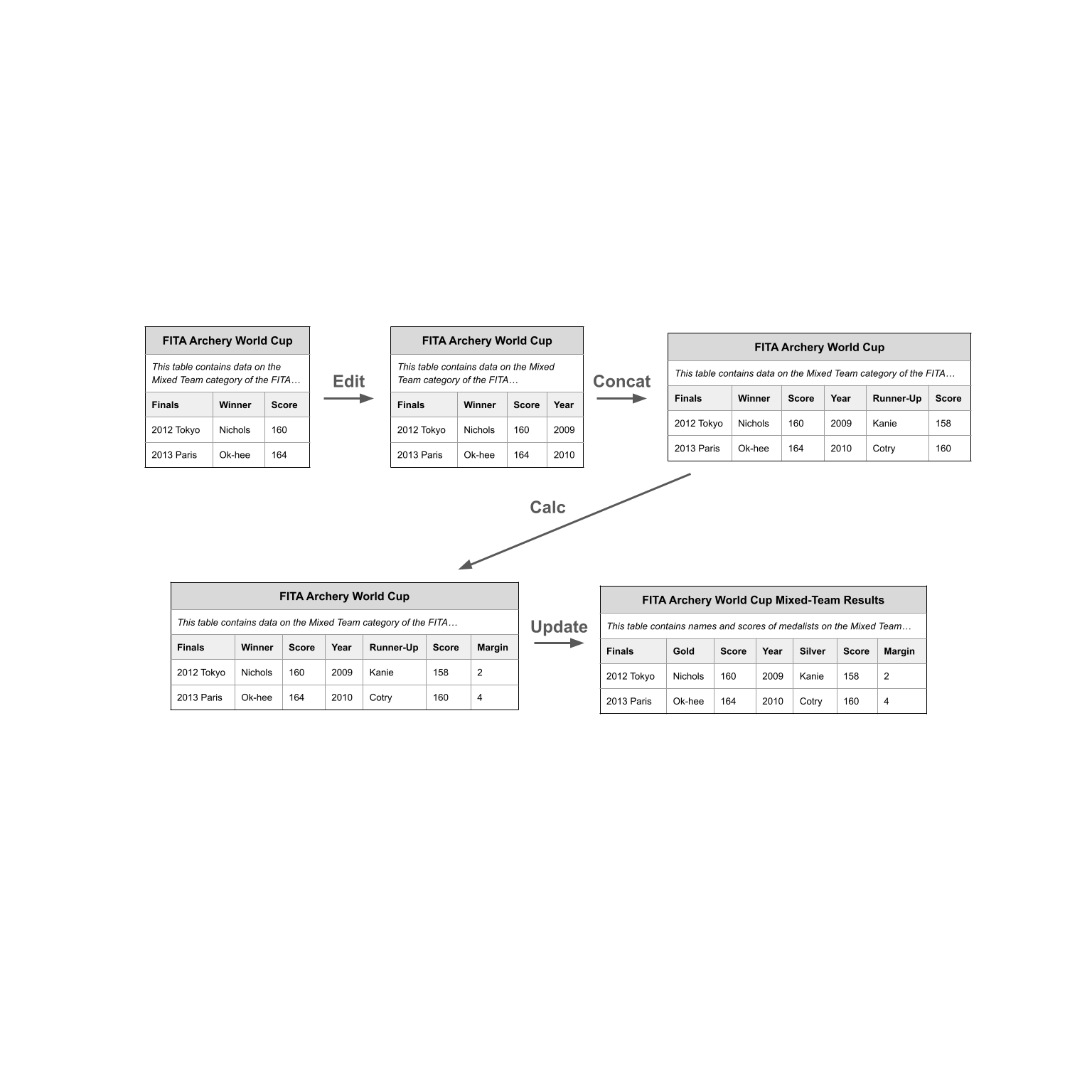}
    \caption{Four tabular transformations performed on a sample table.}
    \label{fig:operations}
\end{figure}

We defined six operations to exemplify the tabular transformations that a data analyst performs on data on the job. Four of these operations rely on an LLM to automatically transform the table: edit, concatenation, calculation, and update. Figure~\ref{fig:operations} shows how these operations may look on a sample table.

The removal operation is applied only to tables without numerical columns, as there may be cases where a table has only one numerical column, and removing it would prevent a calculation process from being executed. The edit operation is applied only to non-numerical tables (those without any numerical columns), while the calculation operation is used exclusively for numerical tables (those with at least one numerical column).

Our pipeline consists of some combination of these six operations performed in the following order: removal, concatenation, edit, calculation, reordering, and update. This ensures that a newly-created column does not subsequently get removed and that tabular transformations are performed before updating any verbiage.

\subsection{Prompt Design}
\label{sec:prompts}

This section details the prompts we used for creating the synthetic data generation pipeline.

The system prompt for asking an LLM to accomplish each operation:
\begin{verbatim}
SYSTEM_PROMPT = "You are a data scientist/analyst who edits tabular data every day."
\end{verbatim}

Prompts for guiding an LLM to accomplish operations: concatenation (concat), edit, calculation (calc), and update:

\begin{verbatim}
CONCAT_OPERATION =  "Make up two new columns with reasonable and diverse values. 
Specifically, each row in cell data should have one more element, 
and the length of column names should increase by one. You can make up data as long
as the values look reasonable."

EDIT_OPERATION = "Create a new column completely based on one or more
existing columns. Some options are but not limited to: binning, string 
operation based on regular expression, information extraction, information 
refinement, etc. After the operation, each row in cell data should have 
one more element, and the length of column names should increase by one."

CALC_OPERATION = "Create a new column completely based on one or more 
existing numerical columns using a type of calculation (mathematical 
calculations, aggregations, allocations, etc.). After the calculation, 
each row in cell data should have one more element, and the length of column 
names should increase by one."

UPDATE_OPERATION = "Update title, column names, and description to match the 
updated cell data."


\end{verbatim}

Finally, here is the prompt that incorporates a serialized table and an operation prompt.

\begin{verbatim}
  {table_data_serialized}

  Your mission is to edit the json-formated tabular datapoint shown above 
  and output the modified table in the exact same format.

  Edit the tabular data with following operations:
  
  {operations_prompt}


  Your output must only be a JSON object, do not explain yourself or output 
  anything else. Again, do not explain yourself or output anything else.

  """
\end{verbatim}

\label{sec:operations}

\end{document}